%% file: main.tex
\documentclass[runningheads]{llncs}

\usepackage{booktabs} 
\usepackage{multirow}

\usepackage[caption=false]{subfig} 

\usepackage{graphicx}

\graphicspath{ {img/} }

\def\Fig#1{Fig.~\ref{fig:#1}}

\def\Tab#1{Tab.~\ref{tab:#1}}

\usepackage{xspace}

\def\CM{\textbf{C1}\xspace}
\def\CP{\textbf{C2}\xspace}
\def\CO{\textbf{C3}\xspace}
\def\CA{\textbf{C4}\xspace}
\def\CB{\textbf{Binary}\xspace}

\usepackage[usenames, dvipsnames]{color}

\begin{document}
\title{Leveraging Medical Visual Question Answering with Supporting Facts}
\titlerunning{Supporting Facts Network}

\author{Tomasz Kornuta \and 
Deepta Rajan \and
Chaitanya Shivade \and \\
Alexis Asseman \and
Ahmet S. Ozcan
}
\authorrunning{T. Kornuta et al.}
\institute{IBM Research AI, Almaden Research Center, San Jose, USA\\
\email{\{tkornut,drajan,cshivade,asozcan\}@us.ibm.com,alexis.asseman@ibm.com}
}

\maketitle              
\begin{abstract}
In this working notes paper, we describe IBM Research AI (Almaden) team's participation in the ImageCLEF 2019 VQA-Med competition. The challenge consists of four question-answering tasks based on radiology images.  The diversity of imaging modalities, organs and disease types combined with a small imbalanced training set made this a highly complex problem. To overcome these difficulties, we implemented a modular pipeline architecture that utilized transfer learning and multi-task learning. Our findings led to the development of a novel model called Supporting Facts Network (SFN). The main idea behind SFN is to cross-utilize information from upstream tasks to improve the accuracy on harder downstream ones. This approach significantly improved the scores achieved in the validation set (18 point improvement in F-1 score). Finally, we submitted four runs to the competition and were ranked seventh.

\end{abstract}

\input{intro}

\input{dataset_short}

\input{model}

\input{experiments}

\input{summary}

\bibliographystyle{splncs04}
\bibliography{bibliography}


\end{document}

%% file: intro.tex
\section{Introduction}
\label{sec:intro}

In the era of data deluge and powerful computing systems, deriving meaningful insights from heterogeneous information has shown to have tremendous value across industries.
In particular, the promise of deep learning-based computational models~\cite{lecun2015deep}  in accurately predicting diseases has further stirred great interest in adopting automated learning systems in healthcare~\cite{2019_ardila_nat_med}.
A daunting challenge within the realm of healthcare is to efficiently sieve through vast amounts of multi-modal information and reason over them to arrive at a differential diagnosis.
Longitudinal patient records including time-series measurements, text reports and imaging volumes form the basis for doctors to draw conclusive insights.
In practice, radiologists are tasked with reviewing thousands of imaging studies each day, with an average of about three seconds to mark them as anomalous or not, leading to severe eye fatigue \cite{syeda2016medical}.
Moreover, clinical workflows have a sequential nature tending to cause delays in triage situations, where the existence of answers to key questions about a patient's holistic conditions can potentially expedite treatment. Thus, building effective question-answering systems for the medical domain by bringing advancements in machine learning research will be a game changer towards improving patient care. 


Visual Question Answering (VQA)~\cite{malinowski2014multi,antol2015vqa} is a new exciting problem domain, where the system is expected to answer questions expressed in natural language by taking into account the content of the image.
In this paper, we present results of our research on the VQA-Med 2019 dataset \cite{ImageCLEFVQA-Med2019}.
The main challenge here, in comparison to the other recent VQA datasets such as TextVQA~\cite{singh2019towards} or GQA~\cite{hudson2019gqa}, is to deal with scattered, noisy and heavily biased data. Hence, the dataset serves as a great use-case to study challenges encountered in practical clinical scenarios.

In order to address the data issues, we designed a new model called \textbf{Supporting Facts Network} (SFN) that efficiently shares knowledge between upstream and downstream tasks through the use of a pre-trained multi-task solver in combination with task-specific solvers. 
Note that posing the VQA-Med challenge as a multi-task learning problem~\cite{caruana1997multitask} allowed the model to effectively leverage and encode relevant domain knowledge.
Our multi-task SFN model outperforms the single task baseline by better adapting to label distribution shifts.




%
%

%% file: dataset_short.tex
\section{The VQA-Med dataset}
\label{sec:dataset}

The VQA-Med 2019~\cite{ImageCLEFVQA-Med2019} is a Visual Question Answering (VQA) dataset embedded in the medical domain, with a focus on radiology images. It consists of: 
\begin{itemize}
\item a training set of 3,200 images with 12,792 Question-Answer (QA) pairs,
\item a validation set of 500 images with 2,000 QA pairs, and
\item a test set of 500 images with 500 questions (answers were released after the end of the VQA-Med 2019 challenge).
\end{itemize}
In all splits the samples were divided into four categories, depending on the main task to be solved:
\begin{itemize}
\item \CM: determine the modality of the image, 
\item \CP: determine the plane of the image,
\item \CO: identify the organ/anatomy of interest in the image, and
\item \CA: identify the abnormality in the image. 
\end{itemize}
Our analysis of the dataset (distribution of questions, answers, word vocabularies, categories and image sizes) led to the following findings and system-design related decisions:
\begin{itemize}
\item merge of the original training and validation sets, shuffle and re-sample new training and validation sets with a proportion of 19:1,
\item use of weighted random sampling during batch preparation,
\item addition of a fifth \CB category for samples with Y/N type questions,
\item focus on accuracy-related metrics instead of the BLEU score,
\item avoid label (answer classes) unification and cleansing,
\item consider \CA as a downstream task and exclude it from the pre-training of input fusion modules,
\item utilization of image size as an additional input cue to the system.
\end{itemize}

%% file: model.tex
\section{Supporting Facts Network}
\label{sec:model}

\begin{figure}
\centerline{\includegraphics[width=0.8\textwidth]{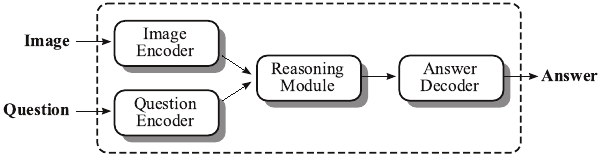}}
\caption{General architecture of Visual Question Answering systems}
\label{fig:vqa_abstract_scheme}
\end{figure}

Typical VQA systems process two types of input, visual (image) and language (question), that need to undergo various transformations to produce the answer.
\Fig{vqa_abstract_scheme} presents a general architecture of such systems, indicating four major modules:  two encoders responsible for encoding raw inputs to more useful representations, followed by a reasoning module that combines them and finally, an answer decoder that produces the answer.

In the early prototypes of VQA systems, reasoning modules were rather simple and relied mainly on multi-modal fusion mechanisms. These fusion techniques varied from concatenation of image and question representations, to more complex pooling mechanisms such as Multi-modal Compact Bilinear pooling (MCB)~\cite{fukui2016multimodal} and Multi-modal Low-rank Bilinear pooling (MLB)~\cite{kim2017hadamard}. Further, diverse attention mechanisms such as question-driven attention over image features~\cite{kazemi2017show} were also used.
More recently, researchers have focused on complex multi-step reasoning mechanisms such as Relational Networks~\cite{santoro2017simple,desta2018object} and Memory, Attention and Composition (MAC) networks~\cite{hudson2018compositional,marois2018transfer}.
Despite that, certain empirical studies indicate early fusion of language and vision signals significantly boosts the overall performance of VQA systems~\cite{malinowski2018visual}.
Therefore, we explored the finding of an "optimal" module for early fusion of multi-modal inputs.

\subsection{Architecture of the Input Fusion Module}
One of our findings from analyzing the dataset was to use the image size as additional input cue to the system. 
This insight triggered an extensive architecture search that included, among others, comparison and training of models with:
\begin{itemize}
\item different methods for question encoding, from 1-hot encoding with Bag-of-Words to different word embeddings combined with various types of recurrent neural networks,
\item different image encoders, from simple networks containing few convolutional layers trained from scratch to fine-tuning of selected state-of-the-art models pre-trained on ImageNet,
\item various data fusion techniques as mentioned in the previous section.
\end{itemize}

\begin{figure}
\subfloat[\textbf{Input Fusion}]{%
	\includegraphics[width=0.61\textwidth]{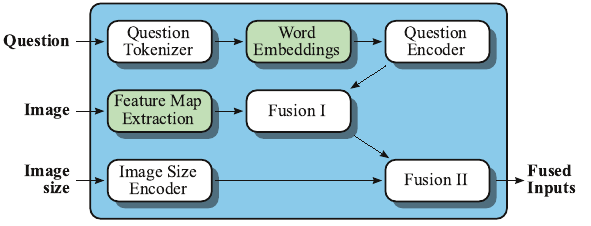}
    \label{fig:vqa_med_input_fusion}
}
\hskip -1cm
\subfloat[\textbf{Question Categorizer}]{%
	\includegraphics[width=0.47\textwidth]{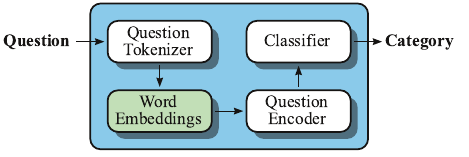}
	\label{fig:vqa_med_question_categorizer}
}
\caption{Architectures of two modules used in the final system}
\end{figure}

The final architecture of our model is presented in \Fig{vqa_med_input_fusion}.
We used GloVe word embeddings~\cite{pennington2014glove} followed by Long Short-Term Memory (LSTM)~\cite{hochreiter1997long}.
The LSTM outputs along with feature maps extracted from images using VGG-16~\cite{simonyan2014very} were passed to the \textbf{Fusion I} module, implementing question-driven attention over image features~\cite{kazemi2017show}.
Next, the output of that module was concatenated in the \textbf{Fusion II} module with image size representation created by passing image width and height through a fully connected (FC) layer.

Note that the green colored modules were initially pre-trained on external datasets (ImageNet and 6B tokens from Wikipedia 2014 and Gigaword 5 datasets for VGG-16 and GloVe models respectively) and later fine-tuned during training on the VQA-Med dataset.

\subsection{Architectures of the Reasoning Modules}

\begin{figure}
\centerline{\includegraphics[width=0.65\textwidth]{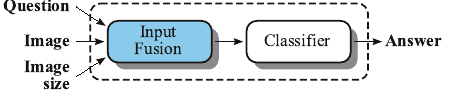}}
\caption{Architecture with a single classifier (IF-1C)}
\label{fig:vqa_med_classification_single_task}
\end{figure}

During the architecture search of the \textbf{Input Fusion} module we used the model presented in \Fig{vqa_med_classification_single_task}, with a simple classifier with two FC layers.
These models were trained and validated on \CM, \CP and \CO categories separately, while excluding \CA.
In fact, to test our hypothesis we trained some early prototypes only on samples from \CA and the models failed to converge.

After establishing the \textbf{Input Fusion} module we trained it on samples from \CM, \CP and \CO categories.
This served as a starting point for training more complex reasoning modules.
At first, we worked on a model that exploited information about $5$ categories of questions by employing $5$ separate classifiers which used data produced by the \textbf{Input Fusion} module.
Each of these classifiers essentially specialized in one question category and had its own answer label dictionary and associated loss function.
The predictions were then fed to the \textbf{Answer Fusion} module, which selected answers from the right classifier based on the question category predicted by the \textbf{Question Categorizer} module, whose architecture is shown in \Fig{vqa_med_question_categorizer}.
Please note that we pre-trained the module in advance on all samples from all categories and froze its weights during the training of classifiers.


\begin{figure}
\centerline{\includegraphics[width=\textwidth]{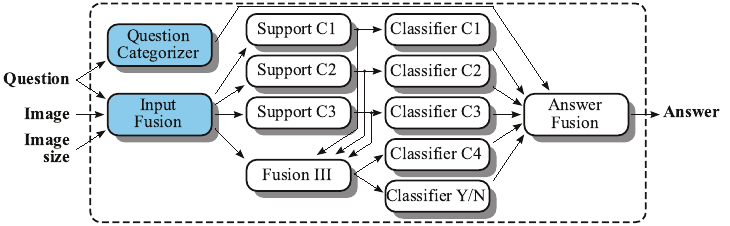}}
\caption{Final architecture of the Supporting Facts Network (SFN)}
\label{fig:vqa_med_classification_five_tasks_supporting_facts}
\end{figure}

The architecture of our final model, \textbf{Supporting Facts Network} is presented in \Fig{vqa_med_classification_five_tasks_supporting_facts}.
The main idea here resulted from the analysis of questions about the presence of abnormalities -- to answer which the system required knowledge on image modality and/or organ type.
Therefore, we divided the classification modules into two networks: Support networks (consisting of two FC layers) and final classifiers (being single FC layers).
We added Plane (\CP) as an additional supporting fact.
The supporting facts were then concatenated with output from \textbf{Input Fusion} module in \textbf{Fusion III} and passed as input to the classifier specialized on \CA questions.
In addition, since \CB Y/N questions were present in both \CM and \CA categories, we followed a similar approach for that classifier.

%% file: experiments.tex
\section{Experimental Results}
\label{sec:experiments}


All experiments were conducted using PyTorchPipe~\cite{ptp2019}, a framework that facilitates development of multi-modal pipelines built on top of PyTorch~\cite{paszke2017automatic}.
Our models were trained using relatively large batches ($256$), dropout ($0.5$) and Adam optimizer~\cite{kingma2014adam} with a small learning rate ($1e-4$).
For each experimental run, we generated a new training and validation set by combining the original sets, shuffling and sampling them in proportions of $19:1$, thereby resulting in a validation set of size $5\%$.

\begin{table}[htbp] 
	\begin{center}
		\begin{tabular}{c@{\extracolsep{9pt}}ccccccccc}
			\toprule
			& \multicolumn{3}{c}{Resampled Valid. Set}  & \multicolumn{3}{c}{Original Train. Set} & \multicolumn{3}{c}{Original Valid. Set} \\
			\cmidrule{2-4} \cmidrule{5-7} \cmidrule{8-10} 
			Model & ~Prec. & Recall & F-1 & ~Prec. & Recall & F-1 & ~Prec. & Recall & F-1 \\
			\midrule
			IF-1C & 0.630	& 0.435	& 0.481	& 0.683	 & 0.497	& 0.545	& 0.690	& 0.499	& 0.548 \\
			SFN & 0.759 & 0.758 & 	0.758 & 0.753 & 0.692 & 0.707 & 0.762 & 0.704 & 0.717 \\
			\bottomrule
		\end{tabular}
	\end{center}
	\caption{Summary of experimental results. All columns contain average scores achieved by $5$ separately trained models on resampled training and validation sets. We also present scores achieved by the models on original sets (in the evaluation mode).}
	\label{tab:results}
\end{table}

In \Tab{results} we present a comparison of average scores achieved by our baseline models using single classifier (IF-1C) and the Supporting Facts Networks (SFN).
Our results clearly indicate the advantage of using \textit{'supporting facts'} over the baseline model with a single classifier.
The SFN model by our team achieved a best score of ($0.558$ Accuracy, $0.582$ BLEU score) on the test set as indicated by the CrowdAI leaderboard.
One of the reasons for such a significant drop in performance is due to the presence of new answers classes in the test set that were not present both in the original training and validation sets.

%% file: summary.tex
\section{Summary}
\label{sec:summary}

	
In this work, we introduced a new model called \textbf{Supporting Facts Network} (SFN), that leverages knowledge learned from combinations of upstream tasks in order to benefit additional downstream tasks. 
The model incorporates domain knowledge that we gathered from a thorough analysis of the dataset, resulting in specialized input fusion methods and five separate, category-specific classifiers.
It comprises of two pre-trained shared modules followed by a reasoning module jointly trained with five classifiers using the multi-task learning approach.
Our models were found to train faster and to deal much better with label distribution shifts under a small imbalanced data regime.

Among the five categories of samples present in the VQA-Med dataset, \CA and \CB turned out to be extremely difficult to learn, for several reasons.
First, there were $1483$ unique answer classes assigned to $3082$ training samples related to \CA.
Second, both \CA and \CB required more complex reasoning and, besides, might be impossible to conclude by looking only at the question and content of the image.
However, our observation that some of the information from simpler categories might be useful during reasoning on more complex ones, we refined the model by adding supporting networks.
Given, modality, imaging plane and organ typically help narrow down the scope of disease conditions and/or answer whether or not an abnormality is present.
Our empirical studies prove that this approach performs significantly better, leading to an 18 point improvement in F-1 score over the baseline model on the original validation set.
